\documentclass[10pt,journal,final]{IEEEtran}

\usepackage{times}
\usepackage{epsfig}
\usepackage{graphicx}
\usepackage{amsmath}
\usepackage{amssymb}
\usepackage{multirow}
\usepackage{color}
\usepackage{algorithm,algorithmic}
\usepackage{lettrine}

\ifCLASSINFOpdf

\else

\fi

\begin{document}

\title{Complete Scene Reconstruction by Merging Images and Laser Scans}

\author{Xiang~Gao, Shuhan~Shen, Lingjie~Zhu, Tianxin~Shi, Zhiheng~Wang, and~Zhanyi~Hu
\thanks{Manuscript received January 21, 2019; revised May 16, 2019 and August 6, 2019; accepted September 15, 2019. This work was partially supported by the National Science Foundation of China (NSFC) under grants 61632003 and 61873265, partially supported by Henan Science and Technology Innovation Outstanding Youth Program under grant 184100510009, and partially supported by Henan University Scientific and Technological Innovation Team Support Program under grant 19IRTSTHN012. \emph{(Corresponding author: Shuhan Shen.)}}
\thanks{X. Gao is with National Laboratory of Pattern Recognition, Institute of Automation, Chinese Academy of Sciences, Beijing 100190, China, and is also with College of Engineering, Ocean University of China, Qingdao 266100, China (e-mail: xiang.gao@nlpr.ia.ac.cn).}
\thanks{S. Shen, L. Zhu, T. Shi, and Z. Hu are with National Laboratory of Pattern Recognition, Institute of Automation, Chinese Academy of Sciences, Beijing 100190, China, and are also with University of Chinese Academy of Sciences, Beijing 100049, China (e-mail: \{shshen, lingjie.zhu, tianxin.shi, huzy\}@nlpr.ia.ac.cn).}
\thanks{Z. Wang is with College of Computer Science and Technology, Henan Polytechnic University, Jiaozuo 454000, China (e-mail: wzhenry@eyou.com).}
\thanks{Copyright \copyright 2019 IEEE. Personal use of this material is permitted. However, permission to use this material for any other purposes must be obtained from the IEEE by sending an email to pubs-permissions@ieee.org.}}

\markboth{IEEE Transactions on Circuits and Systems for Video Technology,~Vol.~XX, No.~XX, Month~Year}
{Gao \MakeLowercase{\textit{et al.}}: Complete Scene Reconstruction by Merging Images and Laser Scans}

\maketitle

\begin{abstract}
Image based modeling and laser scanning are two commonly used approaches in large-scale architectural scene reconstruction nowadays. In order to generate a complete scene reconstruction, an effective way is to completely cover the scene using ground and aerial images, supplemented by laser scanning on certain regions with low textures and complicated structures. Thus, the key issue is to accurately calibrate cameras and register laser scans in a unified framework. To this end, we proposed a three-step pipeline for complete scene reconstruction by merging images and laser scans. First, images are captured around the architecture in a multi-view and multi-scale way and are feed into a structure-from-motion (SfM) pipeline to generate SfM points. Then, based on the SfM result, the laser scanning locations are automatically planned by considering textural richness, structural complexity of the scene and spatial layout of the laser scans. Finally, the images and laser scans are accurately merged in a coarse-to-fine manner. Experimental evaluations on two ancient Chinese architecture datasets demonstrate the effectiveness of our proposed complete scene reconstruction pipeline.
\end{abstract}


\IEEEpeerreviewmaketitle

\section{Introduction}

\IEEEPARstart{T}{HERE} are two key issues in 3D reconstruction of large-scale architectural scenes: accuracy and completeness. Though many scene reconstruction pipelines focus on the issue of accuracy, they pay less attention to the reconstruction completeness. The common pipelines can achieve good reconstruction completeness in scenes with relatively simple structures. However, when the architectural scene is complicated, \emph{e.g.} ancient Chinese architecture, the reconstruction completeness can hardly be guaranteed. In order to reconstruct an accurate and complete 3D model (point cloud or surface mesh) of the large-scale and complicated architectural scene, both global structures and local details of the scene need to be surveyed. Currently, there are two frequently used surveying ways for scene reconstruction, image based \cite{ChatterjeeICCV13,CuiICCV15,CuiTIP15,SchonbergerCVPR16,CuiCVPR17,CuiPR17,FurukawaPAMI10,ShenTIP13,ShenTIP14,UmmenhoferIJCV17,ParkPAMI17} and laser scanning based reconstruction \cite{ZhengSIGGRAPH10,NanSIGGRAPH10,VanegasTVCG12,LiECCV16}. These two ways are complementary in flexibility and accuracy.

The image based reconstruction method is convenient and flexible. The up-to-date image collection equipment is portable and with high resolution, which is appropriate for complete coverage of large-scale scenes. However, the results of existing image based methods heavily depend on several external factors, \emph{e.g.} illumination variation, textural richness and structural complexity. As a result, there are inevitable errors in the image based reconstruction results, especially in the low textured, low lighting, or complicated structured regions.

The laser scanning based reconstruction method possesses high accuracy and is robust to adverse conditions. However, in order to get a complete coverage of large-scale scenes, multi-viewpoint scanning and registration is required. Usually, the laser scans are coarsely registered with the help of man-made targets, which are manually placed in the scene, and are further finely registered by the iterative closest point (ICP) \cite{BeslPAMI92} method. Thus, to achieve a complete scene reconstruction, plenty of laser scans are required, which is time-consuming and inefficient with the currently cumbersome scanning equipment.

In order to generate a complete scene reconstruction by merging images and laser scans, a straightforward way is to treat images and laser scans equally. Specifically, architectural scene models are obtained from these two kinds of data respectively at first and merged together by ground control points (GCPs) \cite{BastoneroAnnals14} or using ICP method \cite{RussoAnnals14,AltuntasArchives15} afterwards. However, this is non-trivial because the point clouds generated from images and laser scans have significant differences in density, accuracy, completeness, \emph{etc.}, which results in inevitable errors in registration. In addition, the laser scanning locations need to be carefully selected to guarantee the scanning overlap for their self-registration.

In this paper, a more effective data collection and scene reconstruction pipeline is proposed, which takes both the data collection efficiency, and the reconstruction accuracy and completeness into consideration. Our pipeline uses images as primacy to completely cover the scene, and uses laser scans as supplement to deal with the low textured, low lighting, or complicated structured regions. It mainly contains three steps: 1)~image capturing, 2)~laser scanning, and 3)~image and laser scan merging. The images are captured to completely cover the scene and to generate structure-from-motion (SfM) points. Based on the SfM result, the laser scanning locations are automatically planned. Finally, the images and laser scans are merged in a coarse-to-fine manner to generate an accurate and complete scene reconstruction. The advantages of this framework are: 1)~Neither overlaps between laser scans nor man-made targets for registration are mandatory as the laser scans are only served as supplements of the images; 2)~By integrating laser scans into the image based reconstruction framework, the reconstruction accuracy and completeness is increased in turn. To our knowledge, we are the first to merge ground and aerial images and terrestrial laser scans for reconstructing accurate and complete outdoor and indoor scenes.

The main contributions of this paper are threefold: 1)~A novel reconstruction pipeline using images as primacy and laser scans as supplement, which takes both the data collection efficiency, and the reconstruction accuracy and completeness into account; 2)~A fully automatic laser scanning location planning algorithm considering textural richness, structural complexity of the scene, and spatial layout of the laser scans; and 3)~A coarse-to-fine image and laser scan merging method, by which an accurate and complete scene reconstruction is generated.

\section{Related Work}

There are three main categories of works related to ours: 1)~image based reconstruction, 2)~laser scanning based reconstruction, and 3)~scene reconstruction using both images and laser scans.

\subsection{Image Based Reconstruction}

Reconstructing scenes from images is the most obvious way as it is the closest way to that of human perceiving the real world. Image based reconstruction has many advantages, \emph{e.g.} easy to obtain, store and distribute, low-cost, and flexible.

The pipeline of image based reconstruction goes as follows. First, feature detection is performed for individual image and feature matching is performed for image pair \cite{LoweIJCV04}. When performing feature matching, usually vocabulary tree \cite{NisterCVPR06,SchonbergerACCV16} is used to index target images with high similarity, and fast library for approximate nearest neighbors (FLANN) \cite{MujaPAMI14} is employed to search approximately nearest feature neighbors. By this way, the efficiency of image matching procedure is largely improved. Then, SfM procedure \cite{CuiICCV15,CuiTIP15,SchonbergerCVPR16,CuiCVPR17} is performed on the pair-wise point matches to estimate the camera poses and triangulate the sparse scene points. Next, multi-view stereo (MVS) \cite{FurukawaPAMI10,ShenTIP13,ShenTIP14} is performed based on registered cameras to get dense point cloud. And finally, image based surface reconstruction \cite{UmmenhoferIJCV17,ParkPAMI17} is performed on the point cloud to obtain detailed surface mesh. Though with many advantages, the image based method is vulnerable to illumination variation, low textures and complicated structures. What is more, inevitable mismatching and error accumulation usually lead to scene drifting.

In addition, there are several methods proposed planning camera network either in off-line \cite{HoppeCVWW12,RobertsICCV17} or on-line \cite{HuangICRA18} scheme, which are mainly used for aerial image capturing. These methods focus on how to completely cover the scenes with minimum image overlap and flight time. However, in this paper, we do not seek for the optimal image capturing locations but only try to properly cover the scene with ground and aerial images.

\subsection{Laser Scanning Based Reconstruction}

Compared with the image based reconstruction methods, the laser scanning based ones acquire the scene structures through active vision. As a result, it possesses several advantages, \emph{e.g.} higher accuracy and less dependency on the external circumstances.

However, due to limitation in scanning viewpoint and inconvenience in data collection, the completeness of scene coverage for laser scanning based methods is hard to guarantee. As a result, several methods are proposed achieving a complete scene reconstruction from laser point clouds. Self-similar structures \cite{ZhengSIGGRAPH10} or simple building blocks \cite{NanSIGGRAPH10} are exploited to reconstruct complete scenes (buildings or facades) from incomplete laser scans. Other methods \cite{VanegasTVCG12,LiECCV16} reconstruct scenes from laser scans based on Manhattan-world assumption. Though quite impressive reconstructions could be achieved by these methods above, they either require user interaction \cite{ZhengSIGGRAPH10,NanSIGGRAPH10} or based on strong assumption \cite{VanegasTVCG12,LiECCV16}, which makes their scalability poor.

There are several light detection and ranging (LiDAR) based simultaneous localization and mapping (SLAM) techniques \cite{ZhangRSS14,HessICRA16,ShanIROS18} which obtain laser points of the scenes. By taking the advantages of SLAM techniques and mobile laser scanner, \emph{e.g.} Velodyne LiDAR in \cite{ZhangRSS14,ShanIROS18}, they possess higher efficiency and lower cost compared with the large-scale laser scanner based method. However, these methods only generate laser points with rather low spatial resolution, thus they are not suitable for reconstructing large-scale architectural scenes with complicated structures, especially for the ancient Chinese architecture considered here.

In addition, in order to completely cover the scene with as few laser scans as possible, several methods are proposed dealing with the issue of optimal terrestrial laser scanner network design \cite{SoudarissananeArchives11,WujanzArchives16,JiaAnnals17,JiaAnnals18}. These methods are based on existing 2D building map \cite{SoudarissananeArchives11,JiaAnnals17,JiaAnnals18} or 3D object model \cite{WujanzArchives16}. When performing optimization, several factors are considered. For example, range and incidence angles constraints \cite{SoudarissananeArchives11}, sufficient overlap and surface topography \cite{WujanzArchives16} between laser scans, or multi-scale and hierarchical viewpoint planning \cite{JiaAnnals18}. However, in this paper, laser scans are served as supplements of the images and their locations are planned based on the SfM result. As a result, textural richness and structural complexity of the scene are considered when performing laser scanning location planning here. By this way, accurate and complete reconstruction could be achieved.

\begin{table*}
\centering
\small
\caption{Details of Image Capturing.}
\begin{tabular}{|c|c|c|}
\hline
 & Ground images & Aerial images\\
\hline
Capturing device & A Canon EOS 5D Mark III on a GigaPan Epic Pro & A Sony NEX-5R on a Microdrones MD4-1000\\
\hline
 & $45$ images per image capturing location & $5$ flight paths\\
Capturing mode & Pitch: $-40^{\circ}\sim40^{\circ}$ (step: $20^{\circ}$) & $1$ flight path for nadir images\\
 & Yaw: $0^{\circ}\sim320^{\circ}$ (step: $40^{\circ}$) & $4$ flight paths for $45^{\circ}$ oblique images\\
\hline
Focal length & $35$ mm & $24$ mm\\
\hline
Image resolution & $5760$ px$\times3840$ px & $4912$ px$\times3264$ px\\
\hline
\end{tabular}
\label{tb:image}
\end{table*}

\subsection{Reconstruction Using Images and Laser Scans}

There are several methods reconstructing scenes using both images and laser scans. However, the purposes of involving these two kinds of data are different for different systems.

Some works propose registering 2D images with 3D laser scans by utilizing low level (point or line) \cite{LiuICCV07,BilaAnnals13,SirmacekAnnals13} or high level (plane) \cite{StamosICCV01} features, by which the 3D laser points can be textured from the registered 2D images. Based on the registered 2D images and 3D laser scans, Li \emph{et al.} \cite{LiICCV11} propose fusing images and laser points by leveraging their respective advantages to get a complete, textured, and regularized urban facade reconstruction. In addition, in the communities of photogrammetry \cite{NexAnnals15}, computer vision \cite{StrechaCVPR08,SchopsCVPR17} and computer graphics \cite{KnapitschTOG17}, several benchmarks contain both images and laser scans are proposed for reconstruction method evaluation. However, the laser scans are mostly served as ground truths which are relatively independent to the images. There are several methods \cite{BastoneroAnnals14,RussoAnnals14,AltuntasArchives15} which have similar motivation with ours, \emph{i.e.} integrating images and laser scans for complete scene reconstruction. These methods are based on 3D-3D registration, which is performed using either GCPs \cite{BastoneroAnnals14} or ICP algorithm \cite{RussoAnnals14,AltuntasArchives15}. In comparison, our approach is based on image synthesis and matching. There is no 3D-level large dissimilarity in density, accuracy, and completeness, thus a more accurate merging is achieved by our proposed method.

\section{Proposed Method}

\begin{figure}
\centering
\includegraphics[width=0.486\textwidth]{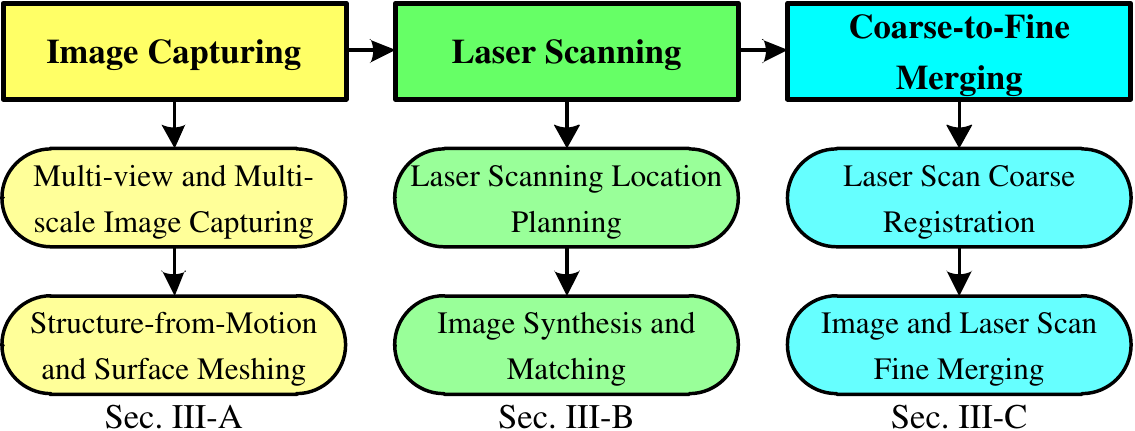}
\caption{Schematic diagram of our proposed complete scene reconstruction pipeline. It mainly contains three steps: 1)~image capturing, 2)~laser scanning, and 3)~coarse-to-fine image and laser scan merging. See text for more details.}
\label{fg:pipeline}
\end{figure}

The pipeline of our proposed complete scene reconstruction method is illustrated in Fig. \ref{fg:pipeline}. Our method mainly contains three steps: 1)~Image capturing. To completely cover large-scale scenes, multi-view and multi-scale image capturing is performed, \emph{i.e.} images are captured from air, ground, outdoor, and indoor. Then, the captured images are matched and feed into a SfM pipeline to generate SfM points. 2)~Laser scanning. Based on the SfM result, laser scanning locations are automatically planned by considering the following three factors: textural richness, structural complexity of the scene and spatial layout of the laser scans. Subsequently, in order to merge images and laser scans, ground-view and aerial-view images are synthesized from the laser points and are matched with the captured images. 3)~Image and laser scan merging. Images and laser scans are merged in a coarse-to-fine manner. The laser scans are coarsely aligned to the SfM points individually at first. Then, images and laser scans are finely merged via a generalized bundle adjustment (BA) with the help of the obtained cross-domain point matches. These three steps are detailed in the following sections respectively.

\subsection{Image Capturing} \label{sc:IC}

\begin{figure}
\centering
\includegraphics[width=0.486\textwidth]{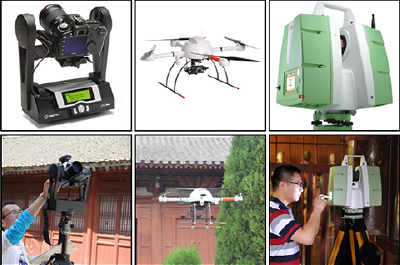}
\caption{Data collection equipments in our experiments. First column: a Canon EOS 5D Mark III mounted on a GigaPan Epic Pro for ground image capturing; Second column: a Sony NEX-5R mounted on a Microdrones MD4-1000 for aerial image capturing; Third column: a Leica ScanStation P30 Scanner for terrestrial laser scanning.}
\label{fg:equip}
\end{figure}

\noindent\textbf{Multi-view and Multi-scale Image Capturing} To ensure the complete coverage of the architectural scenes, in this paper, images are captured in two ways: 1)~close-range ground images with fine connectivity for outdoor and indoor scenes coverage; 2)~large-scale aerial images for entire scene and architectural roof capturing. Some image capturing details of the large-scale ancient Chinese architectural scenes in our experiments are illustrated in Table \ref{tb:image} and Fig. \ref{fg:equip}. The ground (outdoor and indoor) images are captured station by station, which makes it convenient to plan the image capturing locations and efficient to perform the image capturing process. In addition, in order to properly cover the outdoor and indoor scenes from ground viewpoint, the ground image capturing stations are equally spaced in the scene.  Specifically, we coarsely grid the ground plane of the scenes of interest and the grid centers are used as image capturing locations. The side length of the grid is set to $3$ m in this paper. These locations are marked on the ground during the image capturing process, which would be used in the laser scanning step.

\begin{figure}
\centering
\includegraphics[width=0.486\textwidth]{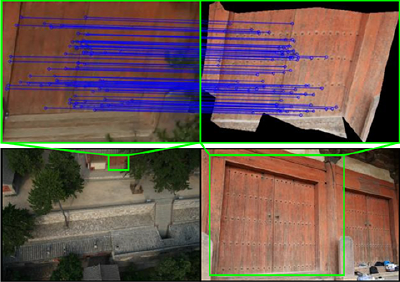}
\caption{An example of ground-to-aerial image feature matching result. First row: point matches between the cropped aerial image (left) and synthetic aerial-view image (right), where the blue segments link the point matches; second row: original aerial and ground image pair with large viewpoint and scale differences.}
\label{fg:match_ag}
\end{figure}

\noindent\textbf{Structure-from-Motion and Surface Meshing} After image capturing, the collected images are feed into a SfM pipeline \cite{CuiCVPR17} to calibrate camera poses and generate spatial points of SIFT features \cite{LoweIJCV04}. In order to merge all captured (outdoor, indoor, and aerial) images into a unified SfM process, ground-to-aerial and outdoor-to-indoor point matches should be generated. However, obtaining these two kinds of point matches are both non-trivial, due to 1)~the large viewpoint and scale differences between ground and aerial images (\emph{cf.} Fig. \ref{fg:match_ag}), and 2)~the limited view overlapping between outdoor and indoor images. In this paper, the scenes captured by aerial images, outdoor images, and indoor images are reconstructed respectively at first, and merged afterwards.

\begin{figure}
\centering
\includegraphics[width=0.486\textwidth]{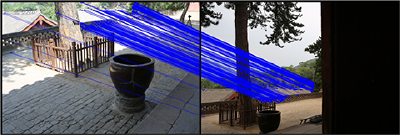}
\caption{An example of outdoor-to-indoor image feature matching result, where the blue segments link the point matches.}
\label{fg:match_oi}
\end{figure}

In recent years, several methods have been proposed integrating ground and aerial data for localization and reconstruction \cite{GaoPRL18}. In this paper, we follow the pipeline proposed in \cite{GaoJPRS18} to merge ground and aerial SfM points. Specifically, for a pair of ground and aerial images, aerial-view image is synthesized from the captured ground image and is matched with the captured aerial image (\emph{cf.} Fig. \ref{fg:match_ag}). The image synthesis is performed by leveraging the co-visible mesh which is generated from ground SfM points using the method \cite{VuPAMI12}. Then, the ground and aerial SfM points are merged via cross-view bundle adjustment. In addition, outdoor and indoor scene merging is also a difficult problem. Recent approach \cite{CohenECCV16} tackles this problem by leveraging the windows, which is not suitable for all building types. \emph{e.g.} ancient Chinese architecture. Here, the outdoor and indoor SfM points are merged with the help of the point matches between the outdoor and indoor images near the doors. The image feature matching result of a pair of outdoor and indoor images near the door is shown in Fig. \ref{fg:match_oi}, which indicates that enough outdoor-to-indoor point matches are generated for outdoor and indoor scene merging. As the image based reconstruction is with scale ambiguity, in order to plan the laser scanning locations and merge the SfM and laser points, the real scale of the merged (outdoor-indoor-aerial) SfM points should be recovered. Here, it is roughly recovered via the built-in GPS of the cameras, by which the SfM points and cameras are geo-referenced. Then, surface reconstruction \cite{VuPAMI12} is performed on the merged SfM points to get surface mesh of the scene, which is used for laser scanning location planning in the following section.

\subsection{Laser Scanning} \label{sc:LS}

\noindent\textbf{Laser Scanning Location Planning} Given the surface mesh reconstructed from the merged SfM points, we try to plan the laser scanning locations with full automation. As the purpose of involving laser scans is to obtain a more accurate and complete scene reconstruction, during scanning location planning, the following three factors should be considered: 1)~textural richness, 2)~structural complexity of the scene, and 3)~spatial distribution of the laser scans. The first two factors mean that the scenes with low textures and complicated structures should be complemented by laser scanning in priority. The third factor means that the laser scanning locations should evenly distribute and not overlap much with each other, in order to save time and cost. Following these three factors, we proposed a method to automatically plan the laser scanning locations.

In order to plan the laser scanning locations, we first obtain several potential laser scanning locations. Then, the scanning location planning becomes a $0$-$1$ integer linear programming problem: Selecting some potential locations as the actual scanning locations (labeled as $1$) and discarding the others (labeled as $0$). The potential laser scanning locations can be simply determined as follows. The ground plane is detected and divided into grids at first. Then, the grid centers are used as the potential scanning locations \cite{SoudarissananeArchives11,JiaAnnals17,JiaAnnals18}. However, in this paper, the potential scanning locations are selected as the capturing locations of the ground images, which has two advantages: 1)~The image capturing locations are carefully selected to properly cover the scene, thus their subset is appropriate for performing laser scanning as well. 2)~During the merging of images and laser scans in the following section, the point matches between the captured ground images and the ground-view images synthesized from the laser scans are required, thus scanning at the image capturing locations benefits the image synthesis and matching procedure.

\begin{figure*}
\centering
\includegraphics[width=0.853\textwidth]{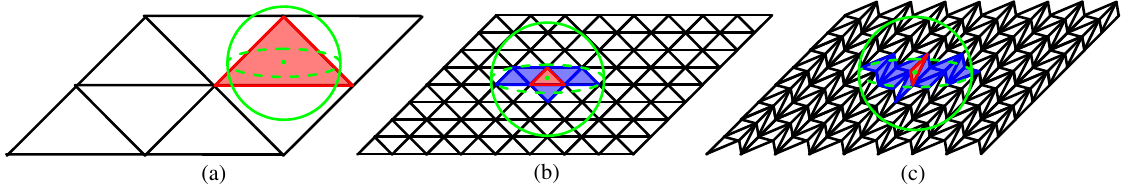}
\caption{A toy example for $a_{i,m}$ computation in three different situations: (a). a region with low texture, (b) a region with rich texture and simple structure, and (c) a region with rich texture and complicated structure. The red facets are those intersected with the rays cast from the potential scanning location. The green balls denote the region for searching neighbour facets of the intersected ones, whose radius is $r_f$. $a_{i,m}$ is computed by summing up the area of the intersected facet (red) and those of its neighbour facets (blue).}
\label{fg:aim}
\end{figure*}

After obtaining the potential laser scanning locations, the actual scanning locations are selected among the potential ones with the help of the surface mesh generated from SfM points. Specifically, we evenly cast $n_r$ rays, $n_r=1000$ in this paper, at each potential scanning location \cite{Deserno04}. The rays casting from the $i$-th location intersect the surface mesh with $n_i$ facets using CGAL\footnote{https://www.cgal.org/}. They are denoted as:
\begin{equation} \label{eq:Fi}
\mathcal{F}_i=\{f_{i,m}\}(i=1,2,\dots,N_p;m=1,2,\dots,n_i),
\end{equation}
where $N_p$ is the number of potential scanning locations, and $f_{i,m}$ (red ones in Fig. \ref{fg:aim}) is the $m$-th intersected facet at the $i$-th location ($n_i$ intersected facets in total). These facets together with their neighbours are used to indicate the textural richness and structural complexity of the scenes around those potential scanning locations. Specifically, for each facet $f_{i,m}$, we obtain its neighbour facets (blue ones in Fig. \ref{fg:aim}) whose distances to $f_{i,m}$ are less than $r_f$, $r_f=0.1$ m in this paper. These neighbour facets are denoted as: $\{n_{i,m,p}\}(p=1,2,\dots,q)$, where $q$ is the number of neighbour facets of $f_{i,m}$. The distance between two facets here is defined by the Euclidean distance between the two facet centers. Then, the areas of $f_{i,m}$ (denoted as $\alpha_{i,m}$) and those of its neighbours (denoted as $\{\beta_{i,m,p}\}$) are summed up and denoted as $a_{i,m}$:
\begin{equation} \label{Eq:aim}
a_{i,m}=\alpha_{i,m}+\sum\limits_{p=1}^{q}\beta_{i,m,p}.
\end{equation}

The value of $a_{i,m}$ is used to indicate the textural richness and structural complexity of the scene near $f_{i,m}$. On the one hand, the textural richness is evaluated based on the assumption that in the regions with lower textures (whether they are with complicated structures or not), less points and larger facets would be generated. Thus, even with few or no neighbour facets, larger $a_{i,m}$ would be obtained due to larger $\alpha_{i,m}$ (\emph{cf.} Fig. \ref{fg:aim}a). On the other hand, the structural complexity is evaluated based on the assumption that in the regions with similar textural richness, more facets are generated to cover the structural complexity for the ones with more complicated structures. Therefore, though with similar facet areas, larger $a_{i,m}$ would be obtained again due to larger $q$ (\emph{cf.} Fig. \ref{fg:aim}b and \ref{fg:aim}c). As a result, the scene near $f_{i,m}$ with lower texture or more complicated structure tends to have larger $a_{i,m}$.

\begin{algorithm}[t]
    \renewcommand{\algorithmicrequire}{\textbf{Input}}
    \renewcommand{\algorithmicensure}{\textbf{Output}}
	\caption{Laser Scanning Location Planning} \label{al:LaserPlan}
	\begin{algorithmic}[1]
		\REQUIRE $A_i$ defined in Eq. \ref{eq:Ai} and $IoU_{i,j}$ defined in Eq. \ref{eq:IoUij}
		\ENSURE Selected indices of the potential scanning locations
        \emph{Initialization:}
		\STATE Select the first scanning location by Eq. \ref{eq:i1}, $N_s\gets1$\\
        \emph{Iteration:}
        \WHILE {the condition defined in Eq. \ref{eq:truncation} satisfies}
        \STATE Select one scanning location by Eq. \ref{eq:nk1}, $N_s\gets N_s+1$
        \ENDWHILE
	\end{algorithmic}
\end{algorithm}

Then, we use
\begin{equation} \label{eq:Ai}
A_i=\frac{\sum\limits_{m=1}^{n_i}a_{i,m}}{n_i},
\end{equation}
where $n_i$ is defined in Eq. \ref{eq:Fi}, to indicate the textural richness and structural complexity of the scene around the $i$-th potential scanning location.

In addition, in order to indicate the overlap between the $i$-th and $j$-th potential scanning location, the intersection over union (IoU) of their intersected facet sets are used, which is denoted as:
\begin{equation} \label{eq:IoUij}
IoU_{i,j}=\frac{\mathcal{F}_i\bigcap\mathcal{F}_j}{\mathcal{F}_i\bigcup\mathcal{F}_j}.
\end{equation}
As the planned laser scanning locations with more even distribution are preferred, the potential locations with less IoUs between each other should be selected in priority. As a result, we formulate the problem of laser scanning location planning as follows:
\begin{equation} \label{eq:LaserPlan}
\begin{split}
\underset{x_i}\max&\frac{\sum\limits_{i=1}^{N_p}A_ix_i}{\sum\limits_{i=1}^{N_p}\sum\limits_{j=i+1}^{N_p}x_ix_jIoU_{i,j}},\\
s.t.&\bigcup\limits_{i=1}^{N_p}x_i\mathcal{F}_i\Bigg/\bigcup\limits_{i=1}^{N_p}\mathcal{F}_i<t_c
\end{split}
\end{equation}
where $x_i=0,1(i=1,2,\dots,N_p)$ is the optimization variable, $x_i=1$ means the $i$-th potential scanning location is selected, otherwise $x_i=0$; $t_c$ is a threshold to bound the coverage of laser scanning and is set to $1/8$ in this paper.

However, the problem defined in Eq. \ref{eq:LaserPlan} is $0$-$1$ integer linear programming problem, which is NP-hard. In this paper, we approximately solve the optimization problem by a greedy algorithm and select one potential scanning location at a time \cite{SoudarissananeArchives11,JiaAnnals18}. The algorithm is detailed in the following.

\begin{figure}
\centering
\includegraphics[width=0.486\textwidth]{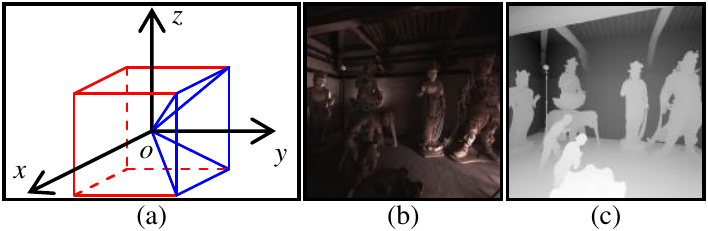}
\caption{(a). Schematic diagram of the virtual cube for ground-view image synthesis, where the blue pyramid denotes one of the virtual cameras. (b). An example of ground-view synthetic image. (c). Depth map of (b).}
\label{fg:cubic}
\end{figure}

\begin{figure}
\centering
\includegraphics[width=0.486\textwidth]{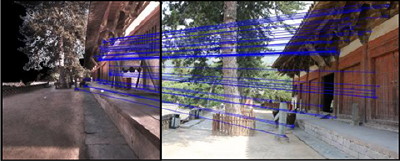}
\caption{An example of synthetic-to-ground image feature matching result, where the blue segments link the point matches.}
\label{fg:match_sg}
\end{figure}

The index of the first selected scanning location is that with largest $A_i$:
\begin{equation} \label{eq:i1}
i_1^*=\arg\max A_i,
\end{equation}
which means the $i_1^*$-th potential scanning location is the first selected one. Suppose after $N_s$ times selection, the indices of the $N_s$ selected potential scanning locations are denoted as $\{i_m^*\}(m=1,2,\dots,N_s)$, while the other $N_p-N_s$ ones are denoted as $\{i_n^{\#}\}(n=1,2,\dots,N_p-N_s)$, which means
\begin{equation}
\begin{split}
&x_{i_1^*}=x_{i_2^*}=\cdots=x_{i_{N_s}^*}=1,\\
&x_{i_1^{\#}}=x_{i_2^{\#}}=\cdots=x_{i_{N_p-N_s}^{\#}}=0.
\end{split}
\end{equation}
Then, during the $N_s+1$-th selection, the $n^*$-th index in $\{i_n^{\#}\}$ is selected as $i_{N_s+1}^*$ by the following optimization:
\begin{equation} \label{eq:nk1}
n^*=\arg\max\frac{\sum\limits_{m=1}^{N_s}A_{i_m^*}+A_{i_n^{\#}}}{\sum\limits_{m=1}^{N_s}\sum\limits_{k=m+1}^{N_s}IoU_{i_m^*,i_k^*}+\sum\limits_{m=1}^{N_s}\sum\limits_{n=1}^{N_p-N_s}IoU_{i_m^*,i_n^{\#}}},
\end{equation}
which means $i_{N_s+1}^*=i_{n^*}^{\#}$. With the selection going on, more and more potential scanning locations are selected to cover the scene. The selection is stopped by the truncation condition in Eq. \ref{eq:LaserPlan}:
\begin{equation} \label{eq:truncation}
\bigcup\limits_{m=1}^{N_s}\mathcal{F}_{i_m^*}\Bigg/\bigcup\limits_{i=1}^{N_p}\mathcal{F}_i<t_c.
\end{equation}
Our proposed automatic laser scanning location planning algorithm is summarized in Algorithm \ref{al:LaserPlan}. Note that laser scanning is performed at the planned locations. As a result, the generated laser points are in the same coordinate system with the geo-referenced SfM points, which makes the following image synthesis and matching procedure straightforward.

\noindent\textbf{Image Synthesis and Matching} After laser scanning location planning, terrestrial laser scanning is performed at those selected scanning locations. In our experiments, a Leica ScanStation P30 Scanner is used (\emph{cf.} Fig. \ref{fg:equip}). Like most up-to-date laser scanners, P30 obtains an extremely large number of accurate spatial points with RGB information. In order to merge image and laser scans, we synthesize images from laser points and match them with the captured ones. In this paper, we not only synthesize the ground-view images, which is similar to that in \cite{SchopsCVPR17}, but also synthesize the aerial-view images. By matching the synthetic and aerial images, more constraints could be obtained for the following image and laser scan merging procedure.

\begin{figure}
\centering
\includegraphics[width=0.486\textwidth]{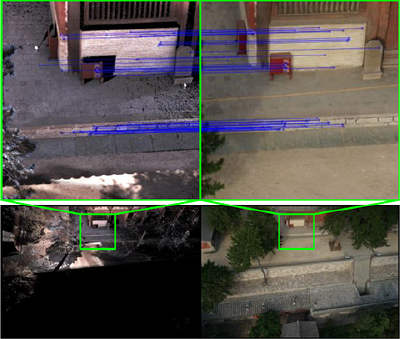}
\caption{An example of synthetic-to-aerial image feature matching result. First row: enlarged synthetic-to-aerial image pair of the green rectangles in the second row to illustrate the feature point matches, which are denoted by the blue segments; second row: original synthetic and aerial image pair.}
\label{fg:match_sa}
\end{figure}

\noindent\emph{Ground-view Image Synthesis} For the points generated from each laser scan, $6$ images are synthesized by projecting them onto the $6$ faces of a virtual cube whose center coincides with the scanning origin (\emph{c.f.} Fig. \ref{fg:cubic}a). The RGBs of synthetic image pixels are the RGBs of the laser points projecting onto them (\emph{c.f.} Fig. \ref{fg:cubic}b). The $6$ cube faces together with the cube center constitute $6$ virtual cameras with orthogonal orientations, which can be seen as a generalized camera model \cite{PlessCVPR03,LeeCVPR13}. Both width and height of the ground-view synthetic image are set to the height of the captured ground image, \emph{i.e.} $3840$ px in this paper.

\noindent\emph{Aerial-view Image Synthesis} We follow the method proposed in \cite{GaoPR18} to select proper aerial images and synthesize images of these selected aerial viewpoints from laser points. For each laser scan, $5$ aerial images are selected to properly cover the laser scan with relatively even distribution. Then, the visible laser points are projected to the selected aerial images with their (intrinsic and extrinsic) camera calibration parameters to synthesize aerial-view images.

As the (ground-view and aerial-view) images are synthesized by point projection, nearest neighbor interpolation is performed to deal with the inevitable missing pixels. In addition, as the 2D-3D correspondences between the synthetic image pixels and the laser points are required in next steps, the depth maps of the synthetic images are generated as well (\emph{c.f.} Fig. \ref{fg:cubic}c). Subsequently, SIFT feature matching between the synthetic and captured images is performed. The ground-view synthetic images are matched with the captured ground images with near locations and similar view directions (less than $5$ m and $45^{\circ}$ respectively in this paper), while the aerial-view synthetic images are matched with the one they are synthesized to. In addition, as the depth near the edge of the synthetic depth map (Fig. \ref{fg:cubic}c) is unreliable, the feature points near the Canny edges \cite{CannyPAMI86} of synthetic depth maps are discarded before image matching. Examples of synthetic-to-ground and synthetic-to-aerial image matching results are given in Fig. \ref{fg:match_sg} and Fig. \ref{fg:match_sa} respectively.

\subsection{Image and Laser Scan Coarse-to-Fine Merging} \label{sc:ILM}

\noindent\textbf{Coarse Registration} The laser scans obtained in Sec. \ref{sc:LS} are coarsely registered to the SfM points one by one as follows. For the $i$-th laser scan, the similarity transformation between its scanning points and the SfM points is denoted as $\{s_i,\boldsymbol{R}_i,\boldsymbol{t}_i\}$. The 3D point correspondences for estimating the similarity transformation are converted from the synthetic-to-ground and synthetic-to-aerial 2D point matches obtained in the last section. The similarity transformation is estimated by a RANSAC-like algorithm \cite{FischlerCACM81}, where a least square solver \cite{UmeyamaPAMI91} is inserted. During the RANSAC procedure in this paper, there are $100$ random samples of the minimal subset ($3$ pairs of 3D point correspondences) and the distance threshold is set to $0.1$ m. The synthetic-to-ground and synthetic-to-aerial 3D point correspondence inliers between the $i$-th laser scan and the SfM points are denoted as $\{\boldsymbol{X}_{i,m}^{GL}\leftrightarrow\boldsymbol{X}_{i,m}^{GI}\}$ and $\{\boldsymbol{X}_{i,n}^{AL}\leftrightarrow\boldsymbol{X}_{i,n}^{AI}\}$ respectively.

\noindent\textbf{Fine Merging} After coarsely registering the laser scans to the SfM points, the (outdoor-indoor-aerial) camera poses, merged SfM points, and the laser scan alignments (similarity transformations) are jointly optimized by a generalized bundle adjustment (BA) to finely merge the images and laser scans. The reasons of performing this further optimization are two-fold: 1)~The SfM points may be not accurate and even with scene drift, especially for the large-scale scenes; 2)~The ground and aerial SfM points may be not accurately merged by the method \cite{GaoJPRS18}. By integrating the SfM result and laser scans into a global optimization, the accuracies of the above two issues are both increased. The BA procedure here is called a generalized one because the camera poses and laser scan alignments are simultaneously optimized by minimizing both 2D-3D reprojection errors and 3D-3D space errors.

The camera poses, merged SfM points, and the laser scan alignments are simultaneously optimized as follows:
\begin{equation} \label{eq:ErEs}
\begin{split}
\underset{\theta}\min\bigg(&\sum_j\sum_k\rho\big(E_R(j,k)\big)\\
+\omega&\sum_i\Big(\sum_m\rho\big(E_S^G(i,m)\big)+\sum_n\rho\big(E_S^A(i,n)\big)\Big)\bigg),
\end{split}
\end{equation}
where $\theta=\{\boldsymbol{R}_j,\boldsymbol{t}_j,\boldsymbol{X}_k,s,\boldsymbol{R}_i,\boldsymbol{t}_i\}$ are the parameters to be optimized. $\boldsymbol{R}_j$ and $\boldsymbol{t}_j$ are the rotation matrix and translation vector of the $j$-th camera; $\boldsymbol{X}_k$ is the $k$-th merged SfM point; $s$ is the uniform scale of all laser scans, and $\boldsymbol{R}_i$ and $\boldsymbol{t}_i$ are the rotation matrix and translation vector of the $i$-th laser scan. The reason of estimating a uniform scale in Eq. \ref{eq:ErEs} lies in that the scale of SfM points recovered via the built-in GPS of the cameras in Sec. \ref{sc:IC} is not accurate enough. In order to achieve a more accurate merging of images and laser scans, the scale ratio between the geo-referenced SfM points and the laser points should be estimated, by which the scale of SfM points can be accurately recovered.

\begin{figure*}
\centering
\includegraphics[width=\textwidth]{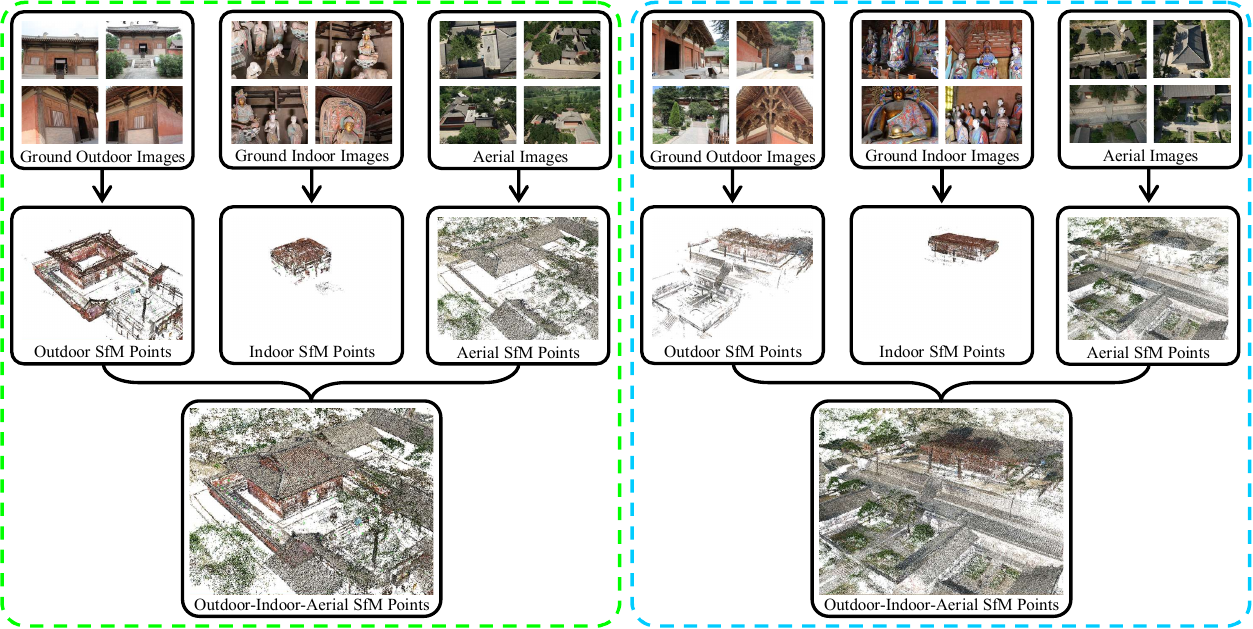}
\caption{Examples of captured images and merged SfM points of NCT (left) and FGT (right).}
\label{fg:sfm}
\end{figure*}

The reprojection error term $E_R(j,k)$ in Eq. \ref{eq:ErEs} is defined as:
\begin{equation} \label{eq:Er}
E_R(j,k)={\Vert\boldsymbol{x}_{j,k}-\gamma(\boldsymbol{K}_j,\boldsymbol{R}_j,\boldsymbol{t}_j,\boldsymbol{X}_k)\Vert}_{\Sigma_{j,k}^{-1}}^2,
\end{equation}
where $\boldsymbol{x}_{j,k}$ is the observed projection of $\boldsymbol{X}_k$ in the $j$-th image; $\boldsymbol{K}_j$ is the intrinsic parameter matrix of the $j$-th camera, which is kept unchanged during the optimization, as it is considered accurately calibrated during the SfM procedure in Sec. \ref{sc:IC}; $\gamma(\cdot)$ is the projection function; $\Sigma_{j,k}$ is the covariance matrix of $\boldsymbol{x}_{j,k}$, which is relevant to the local feature scale of $\boldsymbol{x}_{j,k}$.

The ground space error term $E_S^G(i,m)$ and the aerial spatial error term $E_S^A(i,n)$ in Eq. \ref{eq:ErEs} are respectively defined as:
\begin{equation} \label{eq:Es}
\begin{split}
&E_S^G(i,m)={\Vert s\boldsymbol{R}_i\boldsymbol{X}_{i,m}^{GL}+\boldsymbol{t}_i-\boldsymbol{X}_{i,m}^{GI}\Vert}_{\Sigma_{i,m}^{-1}}^2,\\
&E_S^A(i,n)={\Vert s\boldsymbol{R}_i\boldsymbol{X}_{i,n}^{AL}+\boldsymbol{t}_i-\boldsymbol{X}_{i,n}^{AI}\Vert}_{\Sigma_{i,n}^{-1}}^2,
\end{split}
\end{equation}
where $\Sigma_{i,m}$ and $\Sigma_{i,n}$ are the covariance matrices of $\boldsymbol{X}_{i,m}^{GL}$ and $\boldsymbol{X}_{i,n}^{AL}$ respectively, which are relevant to the distances from the laser points to the scanning origins. The reason of involving Mahalanobis norms in Eq. \ref{eq:Er} and Eq. \ref{eq:Es} is to eliminate the imbalance in dimension and noise level between the reprojection and space error terms.

In addition, $\rho(\cdot)$ in Eq. \ref{eq:ErEs} is the Huber loss function, which is introduced to deal with the inevitable mismatching and noise; and $\omega$ in Eq. \ref{eq:ErEs} is a balancing factor which controls the weights of the constraints defined in Eq. \ref{eq:Er} and Eq. \ref{eq:Es}. The optimization problem in Eq. \ref{eq:ErEs} is solved by Ceres Solver\footnote{http://ceres-solver.org/}. Note that when $\omega\to0$, the optimization problem in Eq. \ref{eq:ErEs} is mainly to minimize the (2D-3D) reprojection errors and approaches a standard BA problem; and when $\omega\to\infty$, the optimization problem is mainly to minimize the (3D-3D) space errors and approaches a laser scan registration problem. A heuristic approach of adaptively setting the value of $\omega$ is described and evaluated in the experimental section.

\section{Experimental Evaluation}

\begin{table}
\centering
\small
\caption{Meta-data of NCT and FGT.}
\begin{tabular}{|c|c|c|}
\hline
Dataset & NCT & FGT\\
\hline
Covering area & $3100$ m$^2$ & $34000$ m$^2$\\
\hline
\# ground outdoor images & $2790$ & $6975$\\
\hline
\# ground indoor images & $855$ & $2475$\\
\hline
Ground outdoor image capturing time & $124$ min & $310$ min\\
\hline
Ground indoor image capturing time & $57$ min & $165$ min\\
\hline
Outdoor-indoor ratio of images & $3.26$ & $2.82$\\
\hline
\# aerial images & $772$ & $1596$\\
\hline
\# planned outdoor laser scans & $6$ & $19$\\
\hline
\# planned indoor laser scans & $5$ & $14$\\
\hline
Outdoor laser scanning time & $180$ min & $570$ min\\
\hline
Indoor laser scanning time & $200$ min & $560$ min\\
\hline
\end{tabular}
\label{tb:data}
\end{table}

In this section, our proposed complete scene reconstruction pipeline is evaluated. We perform experiments on two ancient Chinese architecture datasets, Nan-chan Temple (NCT) and Fo-guang Temple (FGT). They are typical ancient Chinese temple compounds, consisting of one or more main halls and a number of surrounding smaller temples. The indoor scenes of the main halls are usually complicated in structure and low in lighting. As a result, they are suitable objects for the research topic in this paper. We first captured images and generated SfM points as described in Sec. \ref{sc:IC}. Then, we performed laser scanning at the planned scanning locations to obtain laser points using the method in Sec. \ref{sc:LS}. The meta-data of the two datasets is detailed in Table \ref{tb:data}. Note that for ground (outdoor and indoor) images, $45$ images are captured at each capturing location (\emph{cf.} TABLE \ref{tb:image}), which means there are $62(2790\div45)$, $19(855\div45)$, $155(6975\div45)$, and $55(2475\div45)$ image capturing locations for NCT outdoor and indoor scenes, and FGT outdoor and indoor scenes, respectively. In addition, the acquisition time of (outdoor and indoor) ground images and laser points is listed in Table \ref{tb:data}, either. The average acquisition time for ground outdoor and indoor image capturing is about $2$ min and $3$ min per station, while for outdoor and indoor laser scanning is about $30$ min and $40$ min per station. The longer data acquisition time for indoor scenes is due the longer exposure time for scenes with lower lighting.

\subsection{Image Capturing Results} \label{sc:ICR}

We followed the pipeline described in Sec. \ref{sc:IC} to capture images and generate SfM points of NCT and FGT respectively. The number of captured images, including ground outdoor, ground indoor, and aerial ones, for both NCT and FGT is shown in Table \ref{tb:data}. The examples of captured images and reconstructed SfM points are illustrated in Fig \ref{fg:sfm}. We can see from the figure that ground and aerial SfM points are well merged. However, in the regions of low textures, low lighting, or complicated structures of both NCT and FGT, there are only very few points in the merged SfM points. As a result, it is necessary to perform laser scanning to obtain a more accurate and complete architectural scene model.

\subsection{Laser Scanning Results}

\begin{figure}
\centering
\includegraphics[width=0.486\textwidth]{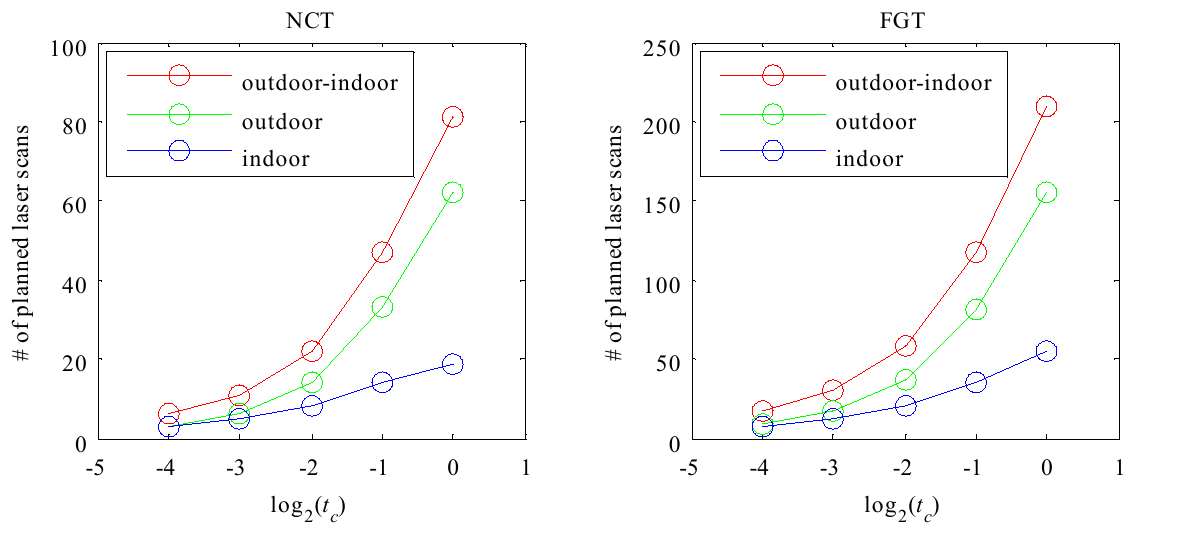}
\caption{The influence of the value of $t_c$ to the number of planned laser scans on NCT and FGT.}
\label{fg:t_c}
\end{figure}

\begin{figure*}
\centering
\includegraphics[width=0.924\textwidth]{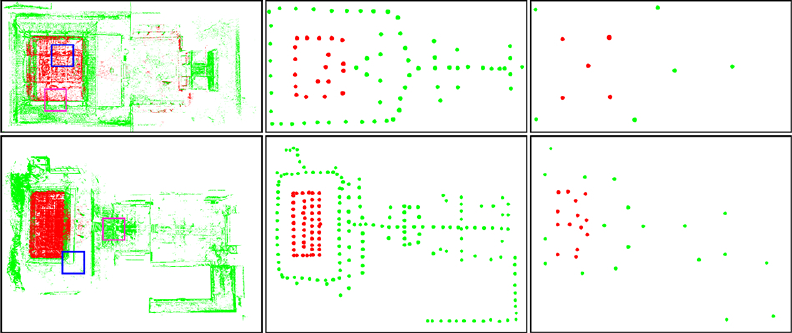}
\caption{Laser scanning location planning results on NCT and FGT. First row: result on NCT; second row: result on FGT. First column: merged ground outdoor (green) and indoor (red) SfM points; second column: outdoor (green) and indoor (red) potential laser scanning locations; third column: outdoor (green) and indoor (red) planned laser scanning locations.}
\label{fg:laser}
\end{figure*}

\begin{figure*}
\centering
\includegraphics[width=0.924\textwidth]{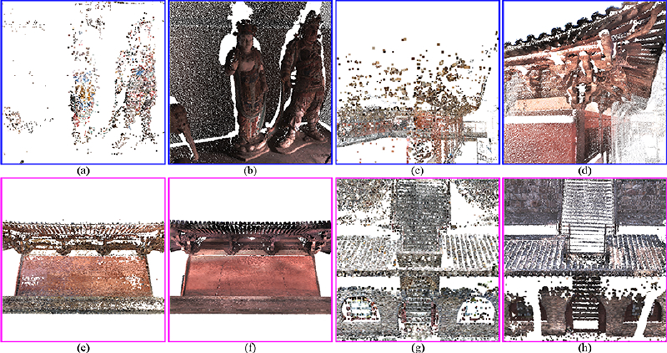}
\caption{(a) and (b). SfM and laser points of the region marked by blue rectangle in top left corner of Fig. \ref{fg:laser}. (c) and (d). SfM and laser points of the region marked by the blue rectangle in bottom left corner of Fig. \ref{fg:laser}. (e) and (f). SfM and laser points of the region marked by magenta rectangle in top left corner of Fig. \ref{fg:laser}. (g) and (h). SfM and laser points of the region marked by the magenta rectangle in bottom left corner of Fig. \ref{fg:laser}.}
\label{fg:sfm_laser}
\end{figure*}

During planning the laser scanning locations by the the method proposed in Sec. \ref{sc:LS}, the parameter $t_c$ in Eq. \ref{eq:LaserPlan} and Eq. \ref{eq:truncation} bounds the coverage of laser scanning, thus it directly determines the number of planned laser scans. Here, we perform experiments on NCT and FGT to demonstrate the influence of the value of $t_c$ ($1/16$, $1/8$, $1/4$, $1/2$, and $1$) on the number of planned laser scans. The results are shown in Fig. \ref{fg:t_c}. From the figure we can see that as the value of $t_c$ getting larger, the number of planned laser scans increases accordingly. In this paper, $t_c$ is set to $1/8$ to balance data collection efficiency and reconstruction results.

\begin{figure*}
\centering
\includegraphics[width=0.930\textwidth]{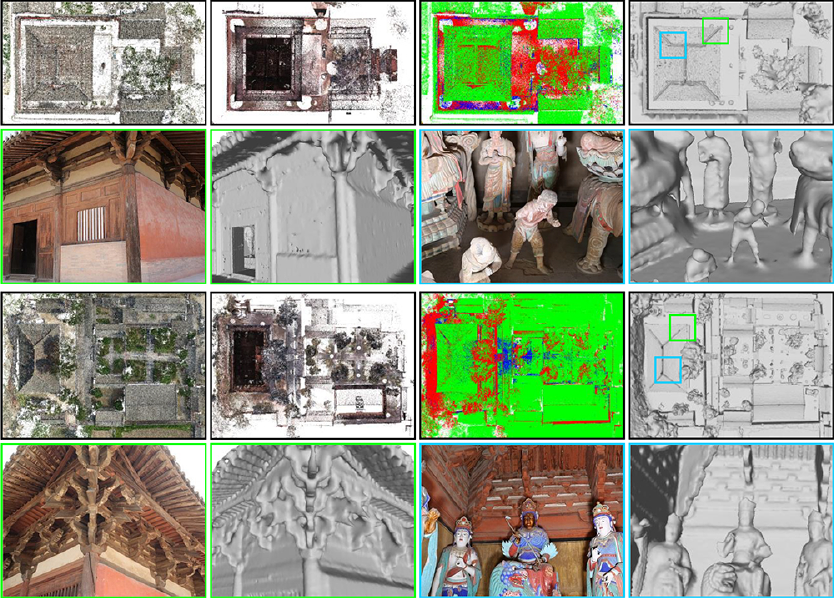}
\caption{Qualitative results of image and laser scan merging on NCT and FGT. First row: long-shots of NCT; from left to right: (outdoor-indoor-aerial) SfM points, (outdoor-indoor) laser points, merged SfM and laser points (red for laser points, green for aerial SfM points, and blue for ground SfM points), surface mesh generated from merged points. Second row: image examples and close-ups of the surface mesh with similar viewpoints on NCT; left two: an outdoor region of the green rectangle at the top right corner of the figure; right two: an indoor region of the rectangle square at the top right corner of the figure. Third and fourth rows: the results on FGT similar to those of the first and second rows.}
\label{fg:merging}
\end{figure*}

By setting $t_c$ to $1/8$, we planned the laser scanning locations and performed scanning. The number of planned (outdoor and indoor) laser scanning locations is given in Table \ref{tb:data}. Note that the outdoor-indoor ratio of images is larger than that of laser scans for both NCT and FGT. As the potential laser scanning locations, \emph{i.e.} the ground image capturing stations, are equally spaced (\emph{cf.} Sec. \ref{sc:IC}), the smaller outdoor-indoor ratio of laser scans means that the density of planned indoor laser scanning locations is larger than that of outdoor ones. That is because compared with the outdoor scenes, the indoor scenes have more complicated structures and lower textures. As a result, relatively more indoor scanning locations are automatically selected from the potential ones by our proposed laser scanning location planning method. Fig. \ref{fg:laser} shows the laser scanning location planning results on NCT and FGT. From the figure we can see that the planned scanning locations are evenly and sparsely distributed throughout the architectural scenes.

In addition, to demonstrate the effectiveness of our proposed laser scanning location planning method, we select two regions for both NCT and FGT, where image based reconstruction method achieves relatively good result in one of them (magenta rectangles in the left column of Fig. \ref{fg:laser}), but not in the other (blue rectangles in the left column of Fig. \ref{fg:laser}). The SfM points and laser points are illustrated in Fig. \ref{fg:sfm_laser}. We can see that there are only a few noisy SfM points at the regions in the first row of Fig. \ref{fg:sfm_laser}. That is because these regions are with low textures (\emph{e.g.} flat walls) and complicated structures (\emph{e.g.} indoor painted sculptures and outdoor bracket sets). As for the regions in the second row of Fig. \ref{fg:sfm_laser}, the SfM points are denser and more accurate due to relatively simple structures and rich textures. As a result, by our proposed laser scanning planning method, the architectural scenes with low textures and complicated structures could be effectively covered by planned laser scans and a more accurate and complete architectural scene model could be obtained.

\subsection{Image and Laser Scan Merging Results} \label{sc:ILMR}

We merged images and laser scans in a coarse-to-fine manner according to the pipeline described in Sec. \ref{sc:ILM}. The qualitative and quantitative merging results are given respectively in the following.

\subsubsection{Qualitative Results}

The qualitative results on NCT and FGT are shown in Fig. \ref{fg:merging}. In order to give a better visualization, we performed surface reconstruction on the merged SfM and laser points using the method \cite{VuPAMI12}. We can see from the long-shots that the images and laser scans are well merged. In addition, the close-ups indicate the accurate and complete scene reconstruction is achieved in the regions with low textures and complicated structures. These qualitative results demonstrate the effectiveness of our proposed image and laser scan merging method.

\begin{table*}
\centering
\small
\caption{Image and laser scan merging accuracies (root-mean-square errors) on NCT and FGT with different ratios of initial space error cost to initial reprojection error cost: $r_c=C_S(\omega)/C_R$.}
\begin{tabular}{|c|c|c|c|c|c|c|c|}
\hline
$\lg(r_c)$ & $-3$ & $-2$ & $-1$ & $0$ & $1$ & $2$ & $3$\\
\hline
NCT & $22.88$ mm & $20.39$ mm & $20.17$ mm & $\boldsymbol{19.42}$ \textbf{mm} & $20.22$ mm & $21.06$ mm & $21.46$ mm\\
\hline
FGT & $33.02$ mm & $32.29$ mm & $28.24$ mm & $\boldsymbol{27.68}$ \textbf{mm} & $30.76$ mm & $35.62$ mm & $39.98$ mm\\
\hline
\end{tabular}
\label{tb:omega}
\end{table*}

\begin{table*}
\centering
\small
\caption{Image and laser scan merging accuracies (root-mean-square errors) on NCT and FGT with different comparative methods.}
\begin{tabular}{|c|c|c|c|c|}
\hline
Method & Baseline: \emph{Coarse} & Knapitsch \emph{et al.} \cite{KnapitschTOG17} & Sch\"{o}ps \emph{et al.} \cite{SchopsCVPR17} & Ours: \emph{Fine}\\
\hline
NCT & $22.78$ mm & $20.79$ mm & $19.88$ mm & $\boldsymbol{19.42}$ \textbf{mm}\\
\hline
FGT & $32.96$ mm & $30.47$ mm & $30.64$ mm & $\boldsymbol{27.68}$ \textbf{mm}\\
\hline
\end{tabular}
\label{tb:comparison}
\end{table*}

\subsubsection{Quantitative Results}

In this section, our proposed image and laser scan merging method is quantitatively evaluated. First, a quantitative measure is introduced for merging accuracy evaluation. Based on the measure, the settings of an important parameter during merging, $\omega$, are assessed; and then the proposed method is quantitatively compared with two state-of-the-arts: Knapitsch \emph{et al.} \cite{KnapitschTOG17} and Sch\"{o}ps \emph{et al.} \cite{SchopsCVPR17}.

\noindent\textbf{Quantitative Measure} As it is difficult to define an exact measure to quantitatively access the merging accuracy, we use an approximate measurement method for the quantitative evaluation. Specifically, we first manually obtain several corresponding points on SfM point cloud and laser point cloud respectively. For both NCT and FGT, $40$ pairs of reference points ($20$ pairs for outdoor and $20$ pairs for indoor), which are evenly distributed in the scenes, are obtained. After image and laser scan merging, each pair of reference point is ideally coincident. Then, the root-mean-square (RMS) value of the distances between all pairs of reference points is used to quantitatively measure the accuracy of image and laser scan merging (the less, the better).

\noindent\textbf{Parameter Settings} Though the imbalance in dimension and noise level between the reprojection and space error terms is eliminated by involving Mahalanobis norm in Eq. \ref{eq:Er} and Eq. \ref{eq:Es}, there is still another imbalance factor, \emph{i.e.} the imbalance in the magnitude of observations, which is caused by the large difference in magnitude between the captured-to-captured image point matches and the synthetic-to-captured image point matches. This imbalance factor influences the image and laser scan merging accuracy largely and in this paper we deal with this issue by involving the balancing factor, $\omega$ in Eq. \ref{eq:ErEs}. Here, we propose an adaptive way of determining the value of $\omega$.

As described in Sec. \ref{sc:ILM}, the optimization problem in Eq. \ref{eq:ErEs} simultaneously optimizes the camera poses and laser scan alignments. Intuitively, when the reprojection error cost approximately equals the space error cost, which means when:
\begin{equation}
\begin{split}
&\sum_j\sum_k\rho\big(E_R(j,k)\big)\\
\approx\omega&\sum_i\Big(\sum_m\rho\big(E_S^G(i,m)\big)+\sum_n\rho\big(E_S^A(i,n)\big)\Big),
\end{split}
\end{equation}
the optimization problem in Eq. \ref{eq:ErEs} achieves a good balance between camera calibration and laser scan registration, and could get a good merging of images and laser scans. To verify this, we respectively define the initial reprojection error cost and the initial space error cost as:
\begin{equation}
\begin{split}
&C_R=\sum_j\sum_k\rho\big(E_R(j,k)\big)\\
&C_S(\omega)=\omega\sum_i\Big(\sum_m\rho\big(E_S^G(i,m)\big)+\sum_n\rho\big(E_S^A(i,n)\big)\Big).
\end{split}
\end{equation}
The initial error cost means the cost is computed from the initial guesses of the parameters to be optimized in Eq. \ref{eq:ErEs}, which are obtained from SfM and coarse laser scan registration processes, thus they are relatively accurate. Let $r_c$ denotes the initial cost ratio $C_S(\omega)/C_R$, and the image and laser scan merging accuracies with different $r_c$ on NCT and FGT are shown in Table \ref{tb:omega}. Note that the value of $r_c$ is proportional to that of $\omega$.

We can see from Table \ref{tb:omega} that for both NCT and FGT, the image and laser scan merging accuracies get higher at first and lower later as $r_c$ gets larger. Only when the value of $r_c$ in a proper range ($\lg(r_c)=-1,0,1$), could the SfM and laser point cloud merging achieves high accuracy, which validates the above assumption. As a result, $\omega$ is set to the value, by which the initial space error cost $C_S(\omega)$ equals the initial reprojection error cost $C_R$ in this paper:
\begin{equation} \label{eq:cost}
\omega=\frac{\sum_j\sum_k\rho\big(E_R(j,k)\big)}{\sum_i\Big(\sum_m\rho\big(E_S^G(i,m)\big)+\sum_n\rho\big(E_S^A(i,n)\big)\Big)}.
\end{equation}

\noindent\textbf{Comparison Results} Then, we quantitatively compared our proposed image and laser scan merging method with Knapitsch \emph{et al.} \cite{KnapitschTOG17} and Sch\"{o}ps \emph{et al.} \cite{SchopsCVPR17}. The comparative results are shown in Table \ref{tb:comparison}. \emph{Coarse} is the merging accuracy after laser scan coarse registration, while \emph{Fine} is the merging accuracy after image and laser scan fine merging. In \cite{KnapitschTOG17}, the dense points generated from images are registered to the laser scans using an extension ICP to similarity transformations (including scale) \cite{LinICIP13}. Note that the merging accuracy of \cite{RussoAnnals14,AltuntasArchives15} would not be higher than that of \cite{KnapitschTOG17}, as their point cloud merging methods are similar in principle. In \cite{SchopsCVPR17}, based on the coarse registration, laser scan alignments are optimized first using point-to-plane ICP \cite{ChenIVC92} and camera poses are then refined by fixing the laser scans using an extended version of the dense image alignment approach \cite{ZhouTOG14}.

\begin{table*}
\centering
\small
\caption{Reconstruction comparison of image based and our methods with different $t_c$ ($1/16$, $1/8$, and $1/4$) against laser scanning based method in precision ($P(\tau)$), recall ($R(\tau)$), and F-score ($F(\tau)$).}
\begin{tabular}{|c|c|c|c|c|}
\hline
Method & Image based method & Our method ($t_c=1/16$) & Our method ($t_c=1/8$) & Our method ($t_c=1/4$)\\
\hline
Measures & $P(\tau)|R(\tau)|F(\tau)$ & $P(\tau)|R(\tau)|F(\tau)$ & $P(\tau)|R(\tau)|F(\tau)$ & $P(\tau)|R(\tau)|F(\tau)$\\
\hline
NCT indoor scene & $95.16|36.21|52.46$ & $95.83|71.58|81,95$ & $97.09|95.47|96.78$ & $97.58|98.24|97.91$\\
\hline
GEH outdoor scene & $93.78|49.29|64.62$ & $94.21|81.61|87.46$ & $96.92|95.07|95.99$ & $97.80|98.72|98.26$\\
\hline
\end{tabular}
\label{tb:reconstruction}
\end{table*}

From the table we can see that compared with the baseline (\emph{Coarse}), the increase in merging accuracy of our method (\emph{Fine}) is larger than both Knapitsch \emph{et al.} \cite{KnapitschTOG17} and Sch\"{o}ps \emph{et al.} \cite{SchopsCVPR17}. That is because: 1)~For \cite{KnapitschTOG17}, as the density and noise level between the points generated from images and laser scans is extremely large, it is hard to achieve an accurate registration between these two kinds of points; 2)~For \cite{SchopsCVPR17}, its merging accuracy is highly dependent on the results of ICP for laser scan alignment optimization. However, the laser scans in our scene reconstruction pipeline are only served as supplement, thus the coverage between each adjacent laser scan pair is quite limited. As a result, for our NCT and FGT datasets, ICP would not achieve a highly accurate registration of the laser scans to help to significantly improve the image and laser scan merging accuracy.

In addition, in order to demonstrate the efficiency advantage in data collection of our proposed pipeline over laser scanning based reconstruction method. We roughly compare the time we used with that of laser scanning based reconstruction method. From Table \ref{tb:data} we can know that the time we used for capturing ground images and scanning laser points on NCT and FGT are about $561(124+57+180+200)$ min and $1605(310+165+570+560)$ min, respectively. However, for laser scanning based reconstruction method, suppose that the coverage threshold $t_c$ is set to $1/2$. As shown in Fig. \ref{fg:t_c}, for NCT, $30$ outdoor and $13$ indoor laser scans are planned; while for FGT, $79$ outdoor and $36$ indoor laser scans are planned. Thus, the time used for laser scanning is about $1420(30\times30+13\times40)$ min and $3810(79\times30+36\times40)$ min on NCT and FGT, respectively. Note that the time for equipment handling, which is much longer for the laser scanner compared with the digital camera, is not included in the reported time above. As a result, our proposed method is much more efficient in data collection compared with laser scanning based method.

\subsection{Scene Reconstruction Results}

Finally, to demonstrate the effectiveness of our scene reconstruction method, we compare it with image based reconstruction method and laser scanning based method. Here, we do not perform the comparison on the whole scenes of NCT and FGT but only on a part of them. For NCT, its indoor scene is used for comparison; while for FGT, the outdoor scene around a hall inside it, named Great East Hall (GEH), is used for comparison.

In order to compare the reconstruction results, the results of laser scanning based method are served as ground truths, and the results of image based method and our method are compared against them. When performing the comparison, the aerial SfM points for the results of both image based method (obtained in Sec. \ref{sc:ICR}) and our method (obtained in Sec. \ref{sc:ILMR}) are eliminated, as there is no airborne laser scanning data.

\begin{figure}
\centering
\includegraphics[width=0.486\textwidth]{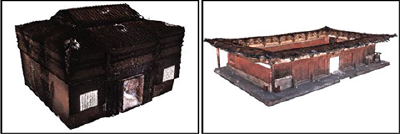}
\caption{Constructed ground truths for indoor scene of NCT (left) and outdoor scene of GEH (right).}
\label{fg:gndtruth}
\end{figure}

To construct ground truths, firstly, several laser scans are additionally performed, which are independent of the ones for scene reconstruction; then, they are accurately registered by using man-made targets and software provided by Leica. The number of additional laser scans for the indoor scene of NCT is $9$ while that for the outdoor scene of GEH is $15$. The constructed ground truths are shown in Fig. \ref{fg:gndtruth}.

After ground truth construction, we follow the method in \cite{KnapitschTOG17} to evaluate the reconstruction results. Specifically, both the reconstructed and ground-truth point clouds are resampled using a uniform voxel grid at first, whose size is $\tau/2$ and $\tau=0.01$ m in this paper. Then, the precision $P(\tau)$, the recall $R(\tau)$, and the comprehensive measure F-score $F(\tau)=\frac{2P(\tau)R(\tau)}{P(\tau)+R(\tau)}$ are computed and served as measures for reconstruction evaluation. The definition of $P(\tau)$ and $R(\tau)$ can be found in \cite{KnapitschTOG17}. The evaluation results based on the above three measures for image based method method and our method with different $t_c$ ($1/16$, $1/8$, and $1/4$) are shown in Table \ref{tb:reconstruction}. We can see from the table that compared with the image based method, our method achieves better performances in both accuracy (precision) and completeness (recall). In addition, for our method, as the value of $t_c$ getting larger, all three measure values are becoming larger accordingly. When $t_c\geq1/8$, the F-score of our method is close to $100$, which means our method achieves comparable reconstruction results in both accuracy and completeness compared with the laser scanning based method. As a result, $1/8$ is an appropriate value for $t_c$ to balance the efficiency of data collection and the performance of scene reconstruction.

\section{Conclusion}

In this paper, we propose a novel pipeline for architectural scene reconstruction by utilizing two different sources of complementary data, images and laser scans, to achieve a good balance between data acquisition efficiency and reconstruction accuracy and completeness. The images are used as primacy to completely cover the scene, while the laser scans are served as supplement to deal with low textured, low lighting, or complicated structured regions. Our pipeline contains three main steps: image capturing, laser scanning, and image and laser scan merging, by which an accurate and complete scene reconstruction is achieved. Experimental results on our two ancient Chinese architecture datasets demonstrate the effectiveness of each main step of our proposed pipeline. In the future, we intend to merge the points scanned from the handy equipment, \emph{e.g.} Kinect, to our pipeline to obtain more complete and detailed reconstruction in complicated architectural scene.

\ifCLASSOPTIONcaptionsoff
  \newpage
\fi

\bibliographystyle{IEEEtran}
\bibliography{bib-file}

\begin{IEEEbiography}[{\includegraphics[width=1in,height=1.25in,clip,keepaspectratio]{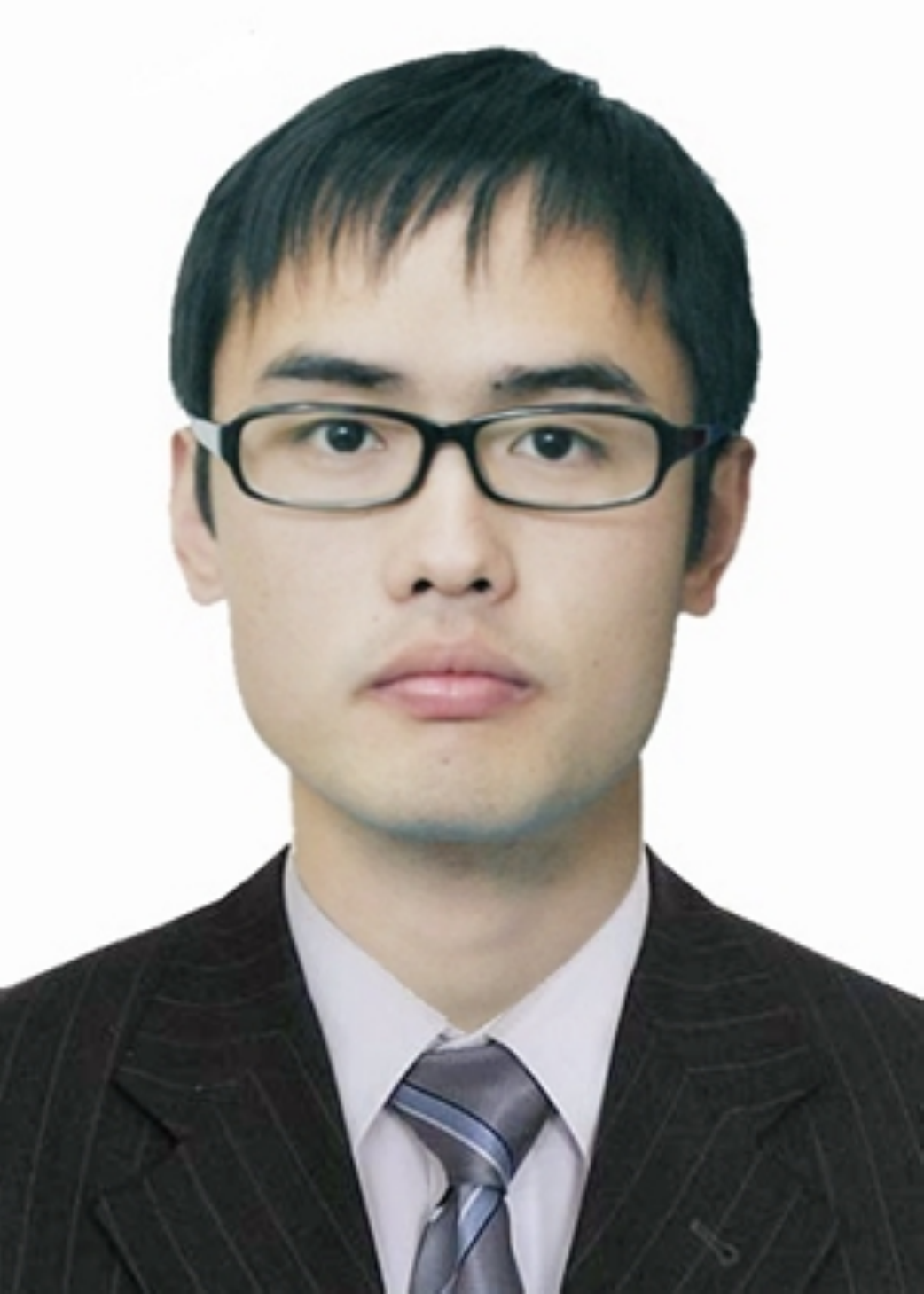}}]{Xiang Gao}
received the B.S. and M.S. Degrees both from Ocean University of China, and the Ph.D. Degree from University of Chinese Academy of Sciences. He is now a lecturer in College of Engineering, Ocean University of China. His research interest is 3D computer vision, with a particular interest in image based large-scale 3D reconstruction.
\end{IEEEbiography}

\begin{IEEEbiography}[{\includegraphics[width=1in,height=1.25in,clip,keepaspectratio]{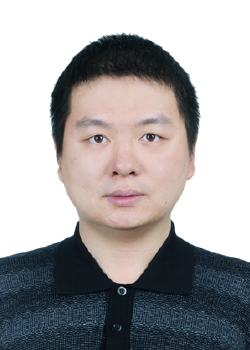}}]{Shuhan Shen}
received the B.S. and M.S. Degrees both from Southwest Jiaotong University, and the Ph.D. Degree from Shanghai Jiaotong University. He is now an associate professor in National Laboratory of Pattern Recognition at Institute of Automation, Chinese Academy of Sciences. His research interests are in 3D computer vision, which include image based 3D modeling of large-scale scenes, 3D perception for intelligent robot, and 3D semantic reconstruction.
\end{IEEEbiography}

\begin{IEEEbiography}[{\includegraphics[width=1in,height=1.25in,clip,keepaspectratio]{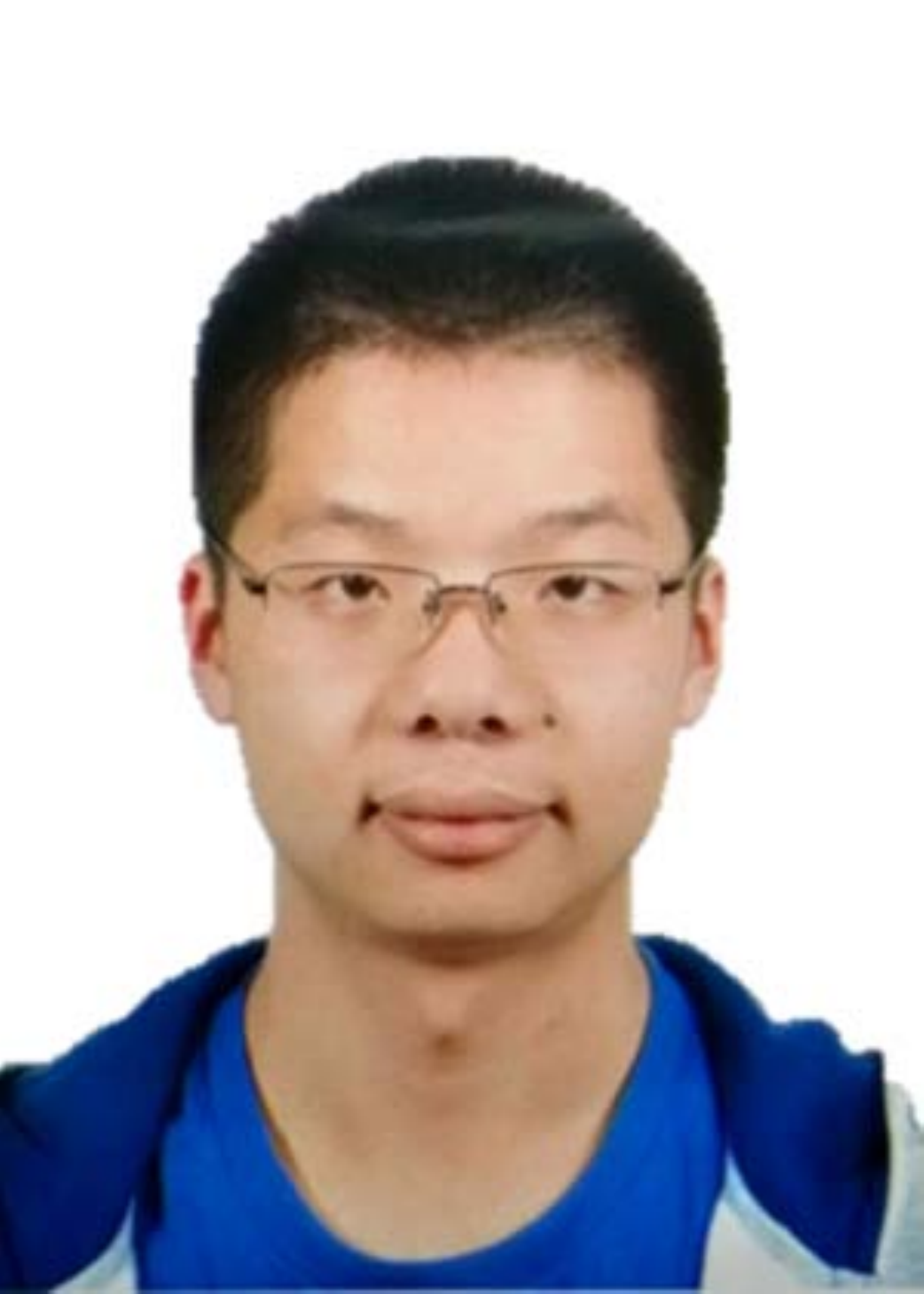}}]{Lingjie Zhu}
received the B.S Degree from Beijing University of Posts and Telecommunications. He is currently a Ph.D. candidate in National Laboratory of Pattern Recognition at Institute of Automation, Chinese Academy of Sciences. His main research interest is 3D computer vision, especially for the procedural reconstruction of urban scenes.
\end{IEEEbiography}

\begin{IEEEbiography}[{\includegraphics[width=1in,height=1.25in,clip,keepaspectratio]{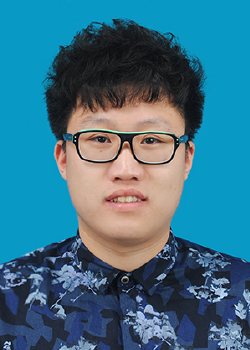}}]{Tianxin Shi}
received the B.S Degree from Beijing University of Posts and Telecommunications. He is currently a M.S. candidate in National Laboratory of Pattern Recognition at Institute of Automation, Chinese Academy of Sciences. His main research interest is 3D computer vision, especially for the visual localization in changing conditions.
\end{IEEEbiography}

\begin{IEEEbiography}[{\includegraphics[width=1in,height=1.25in,clip,keepaspectratio]{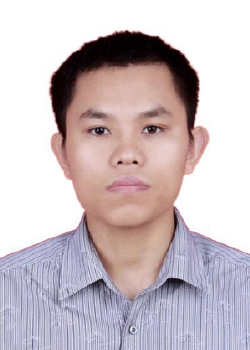}}]{Zhiheng Wang}
received the B.S. Degree from Beijing Institute of Technology, and the Ph.D. Degree from Institute of Automation, Chinese Academy of Sciences. He is currently a Professor in School of Computer Science and Technique, Henan Polytechnic University. His research interests include computer vision, pattern recognition, and image processing.
\end{IEEEbiography}

\begin{IEEEbiography}[{\includegraphics[width=1in,height=1.25in,clip,keepaspectratio]{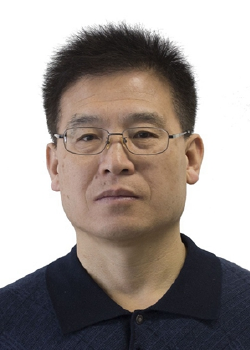}}]{Zhanyi Hu}
received the B.S. Degree from North China University of Technology, and the Ph.D. Degree from University of Liege, Belgium. He is now a professor in National Laboratory of Pattern Recognition at Institute of Automation, Chinese Academy of Sciences, a deputy editor-in-chief for Chinese Journal of CAD and CG, and an associate editor for Science China, and Journal of Computer Science and Technology. He was the organization committee co-chair of ICCV 2005, and a program co-chair of ACCV 2012. He has published more than 150 peer-reviewed journal papers, including IEEE T-PAMI, IEEE T-IP, IJCV, and PR. His current research interests include biology-inspired vision and large-scale 3D scene reconstruction from images.
\end{IEEEbiography}

\end{document}